\documentclass[runningheads]{llncs}
\usepackage{amsmath,amsfonts}
\usepackage{algorithmic}
\usepackage{algorithm}
\usepackage{array}
\usepackage[caption=false,font=normalsize,labelfont=sf,textfont=sf]{subfig}
\usepackage{textcomp}
\usepackage{stfloats}
\usepackage{url}
\usepackage{verbatim}
\usepackage{graphicx}
\usepackage{cite}
\usepackage{caption}
\usepackage{nomencl}
\usepackage{ntheorem}   
\usepackage{lipsum}     
\usepackage{graphicx}
\usepackage{parskip}
\usepackage{adjustbox}
\usepackage{xcolor}
\usepackage{hyperref}

\hypersetup{
	colorlinks=true,
	linkcolor=blue,
	filecolor=green, 
	urlcolor=black,
	citecolor=orange
}

\usepackage{bm}

\DeclareRobustCommand{\uvec}[1]{{%
		\ifcsname uvec#1\endcsname
		\csname uvec#1\endcsname
		\else
		\bm{\mathbf{#1}}%
		\fi
}}
\usepackage{graphicx}
\begin{document}
\title{Design and Motion Analysis of a Reconfigurable Pendulum-Based Rolling Disk Robot with Magnetic Coupling}
\titlerunning{Reconfigurable Pendulum-Based Rolling Disk Robot}
%
\author{Ollie Wiltshire\orcidID{0009-0008-6491-4377} \and 
Seyed Amir Tafrishi\orcidID{0000-0001-9829-3144}}
%
%
\institute{School of Engineering, Cardiff University, Queen's Buildings, The Parade, Cardiff, CF24 3AA  \\
\email{\{wiltshireoj, tafrishisa\}@cardiff.ac.uk}\\
}
\maketitle              
\begin{abstract}
Reconfigurable robots are at the forefront of robotics innovation due to their unmatched versatility and adaptability in addressing various tasks through collaborative operations. This paper explores the design and implementation of a novel pendulum-based magnetic coupling system within a reconfigurable disk robot. Diverging from traditional designs, this system emphasizes enhancing coupling strength while maintaining the compactness of the outer shell. We employ parametric optimization techniques, including magnetic array simulations, to improve coupling performance. Additionally, we conduct a comprehensive analysis of the rolling robot's motion to assess its operational effectiveness in the coupling mechanism. This examination reveals intriguing new motion patterns driven by frictional and sliding effects between the rolling disk modules and the ground. Furthermore, the new setup introduces a novel problem in the area of nonprehensile manipulation.

\end{abstract}

\section{Introduction}
Reconfigurable robots possess multiple units/modules that can interact and alter their shape or configuration through external connections. However, these systems often struggle with limitations in versatility and the ability to achieve movement independently. To address this challenge, there is significant interest in coupling rolling robots as modular components. Rolling robots represent a novel form of mobile robotics \cite{armour2006rolling,tafrishi2019design}, enabling displacement of the entire system while being internally actuated. Nonetheless, the development of coupling mechanisms without external components remains an open challenge in the field of reconfigurable robots \cite{seo2019modular}.

Modular Self-Recongifurable Robot (MSRR) \cite{seo2019modular} is a category of mobile robots. MSRRs are constructed as individual modules that house controllers, sensors, and actuation systems, the individual modules can reconfigure benefiting MSRR-type systems to a diverse set of tasks or traversing unpredictable environments \cite{981854, moubarak2012modular}. The modules interconnect using docking systems, summarised in table \ref{tab:advantages_disadvantages}, including mechanical surfaces (such as locking pins or latches), magnetic interaction, and grippers, including hybrid docking combinations. Reconfiguration takes place autonomously or by manual intervention. MSRRs would prove themselves invaluable in search and rescue, environmental monitoring, and harmful or dangerous environments. Due to their modular nature and reconfigurability \cite{moubarak2012modular,seo2019modular}, allows the robots to adjust their morphology to maximize stability and/or manoeuvrability. A reconfigurable robot's ability to dynamically adapt makes for a very versatile robotic system. Therefore using a modular rolling robot at the the centre of the reconfigurable system could prove to be beneficial and the rolling motion could be an effective option \cite{liang2020freebot}. However, a proper coupling method between each of the rolling modules needs to be carefully considered and is still an open problem.

\begin{table}[t!]
    \centering
    \caption{Advantages and disadvantages of common coupling systems}
    \vspace{0.3cm}
\begin{adjustbox}{width=4 in,height=2.4 in}  
    \begin{tabular}{>{\raggedright\arraybackslash} p{0.35\linewidth}>{\raggedright\arraybackslash} p{0.35\linewidth}>{\raggedright\arraybackslash} p{0.35\linewidth}}
        \hline
        \textbf{Coupling Types} & \textbf{Advantages} & \textbf{Disadvantages} \\
        \hline
        \textbf{Permanent Magnet} - An array of permanent magnets \cite{6696971} & \begin{itemize}
           {\small \item Self-alignment docking capabilities,
            \item Quick attachment \& good redundancy
            \item Possible multiple docking orientations}
        \end{itemize} & \begin{itemize}
            \item Weak connection 
            \item Fixed permanent coupling
            \item Limited coupled modules and weight-dependent
        \end{itemize}  \\
        \hline
        \textbf{Electromagnet} - An electromagnet coil with/without permanent magnets e.g \cite{OCTABOT,liang2020freebot}.& \begin{itemize}
            \item Stronger connection than permanent magnets
            \item Controllable decoupling
        \end{itemize} & \begin{itemize}
            \item High power consumption and dependency
            \item Overheating possibility
            \item Larger internal space and weight \end{itemize} \\
        \hline
        \textbf{Gripper} - A mechanical gripper  e.g \cite{4020359}. & \begin{itemize}
            \item Strong fixed connection
            \item Direct manipulation of orientation
        \end{itemize}  & \begin{itemize}
            \item Single orientation
            \item Constraint on degrees of freedom
            \item Power consumption
        \end{itemize}\\
        \hline
        \textbf{Mechanical Locking} - Mechanisms such as latches, lips and cones e.g \cite{4059322}. & \begin{itemize}
            \item Strong and rigid connection
            \item Impact resistant
            \item Extendable module number
        \end{itemize} &\begin{itemize}
            \item  Limited by orientation/type of coupling 
            \item  Potential alignment/coupling errors
        \end{itemize}    \\
        \hline
        \textbf{Manual Reconfiguration} -  Manually re-configures to different geometries \cite{YaMoR} & \begin{itemize}
            \item Strong and ridged connection 
            \item Fixable with bolts
            \item More complex geometries achievable
        \end{itemize}  & \begin{itemize}
            \item Not self-reconfigurable
            \item Limits in obstacles traverse
            \item No dynamic adaption
        \end{itemize} 
    \end{tabular} 
    \end{adjustbox}
    \label{tab:advantages_disadvantages}
\end{table}

Rolling robots during motion don't remain fixed to the surface they're upon and are subject to slippage, therefore the state of the robot is lost during movement \cite{ruggiero2018nonprehensile}, therefore prediction of the dynamics is required for accurate control. This motion is a form of non-prehensile manipulation. Non-prehensile manipulation is a form of robot manipulation which involves no direct grasp of the object under control \cite{mason1999progress,ruggiero2018nonprehensile} which poses significant control challenges due to various factors. The object manipulated is subject to slippage and friction, leading to unpredictable dynamics during operation. While some research has focused on the manipulation of rolling and balancing disk on disk \cite{6522171}, other studies have sought to simplify the control of a disk on a beam \cite{8613893}. These studies underscore the complexities inherent in non-prehensile manipulation control. Understanding these challenges is paramount for achieving accurate control of the disk modules. Unlike previous approaches that rely on external actuation for balance, the modular disk module introduced in this paper is internally actuated. However, existing literature does not adequately address the complexities arising from the independent rotation of two modular disk bodies, both in mechanism design and control. This paper aims to showcase new motion control challenges posed by modular mobile disk robots for future non-prehensile manipulation problems in more application-based scenarios.
Thus, the contributions of this paper are as follows:
\begin{itemize}
\item Introduction of a novel internalised pendulum-based magnetic coupling mechanism for rolling reconfigurable disk robots.
\item Investigation of magnetic coupling configurations for pendulum-actuated disk modules.
\item Demonstration of the distinctive motion capabilities of the robot through the implementation of tuned scenario-based PD controllers.
\item Presentation and discussion of a new reconfigurable robotic platform poised to address future challenges in non-prehensile manipulation problems.
\end{itemize}

This paper is organised as follows. Section \ref{sec:Design} introduces the robot design and methods used to optimise the magnetic array, detailing the algorithm used to control the robots' motion and the verification method used to check the stability of the PD controller. Section \ref{sec:motion study} presents the potential motion behaviours of the reconfigurable disk robot, analysing the modules independently and together collaboratively. Finally section \ref{sec:conclusions} concludes the findings of the work so far, improvements to be made and intended future works.

\section{Robot Design and Modelling}\label{sec:Design}
This section centres on the design of the rolling reconfigurable disk modules featuring a magnetic coupling mechanism. Initially, the section outlines the design of the modular bodies, it provides an in-depth examination of magnetic configuration design to ensure sufficient coupling strength between multiple disk modules. Additionally, employing a pendulum-actuated model that approximates the modules as a cart-pendulum system for stable module control with feedback control.

\begin{figure}[t!]
    \centering
    \includegraphics[width=0.75\textwidth]{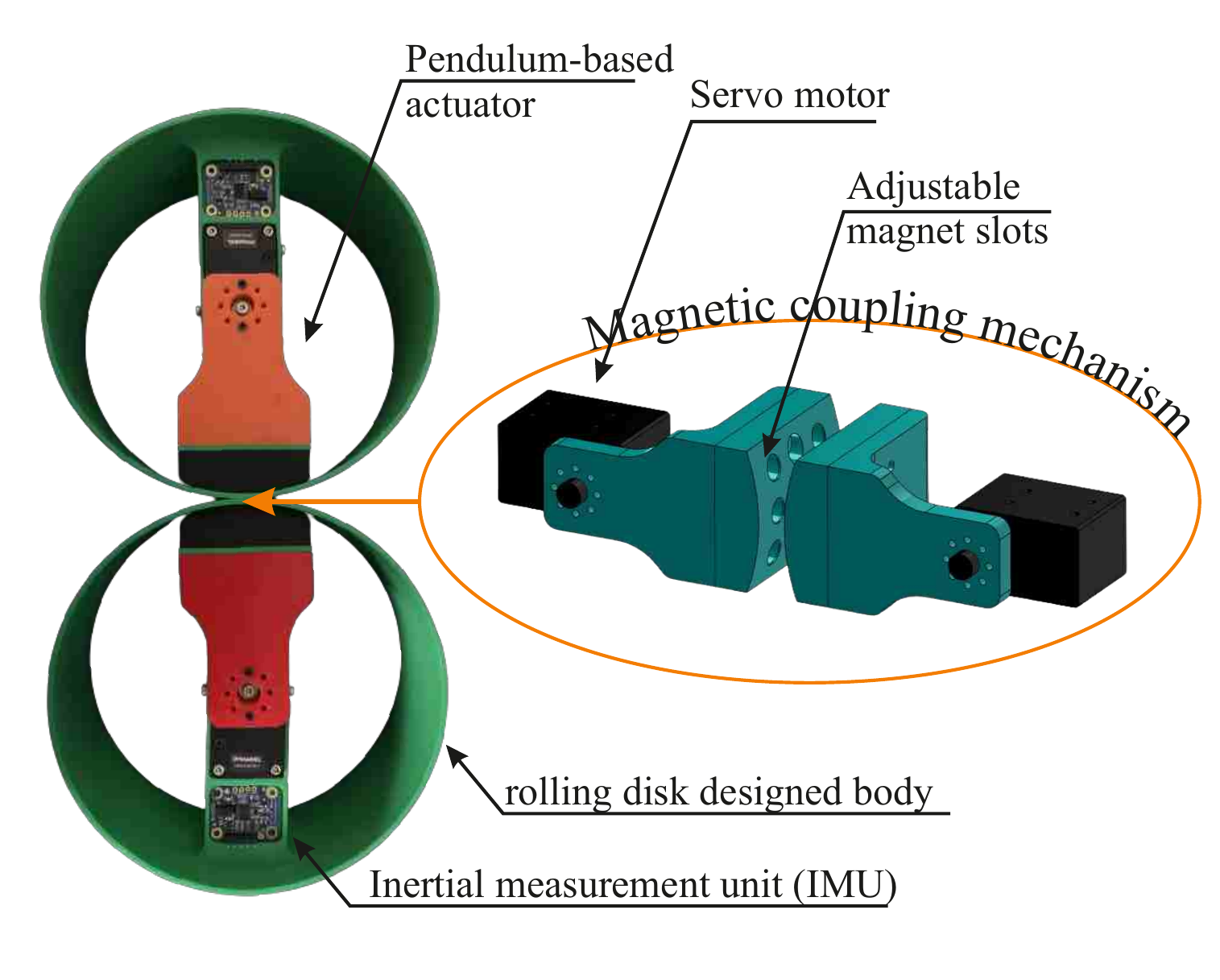}
    \caption{Main body design of reconfigurable rolling disk robot}
    \label{fig: Modular Couple}
    \label{fig:Assembly}
\end{figure}

The main body design of the modular robot features a Dynamixel XM430 high-precision servo motor positioned at the centre, connected to the pendulum-based docking system, as illustrated in Fig. \ref{fig: Modular Couple}. To present the concept of basic behaviours and analyze coupling, the model is simplified to a disk configuration. Here, we consider two modules operating independently of each other, each utilising one motor. Additional bodies can be incorporated by attaching passive magnets in different alignments or by adding extra motors to increase degrees of freedom (DoF). It's important to note that the focus of this study is primarily on the fundamental problem of motion behaviours on coupling and spinning and rolling due to friction, which introduces a new platform for non-prehensile manipulation problem \cite{ruggiero2018nonprehensile}. The robot is controlled with an Arduino MEGA for interfacing with the IMU sensor using a Kalman filter. Meanwhile, the Dynamixel system employs serial bus communication, necessitating conversion via a U2D2 converter for communication with MATLAB which handles the program containing the control loop and data collection.

\begin{figure}[t!]
    \centering
    \includegraphics[width=0.45\linewidth]{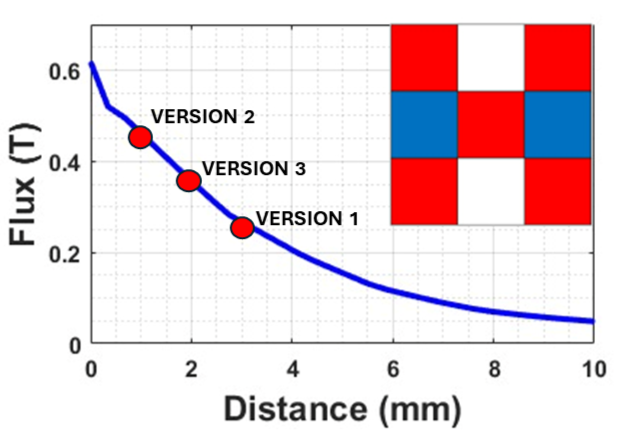}
    \includegraphics[width=0.45\linewidth]{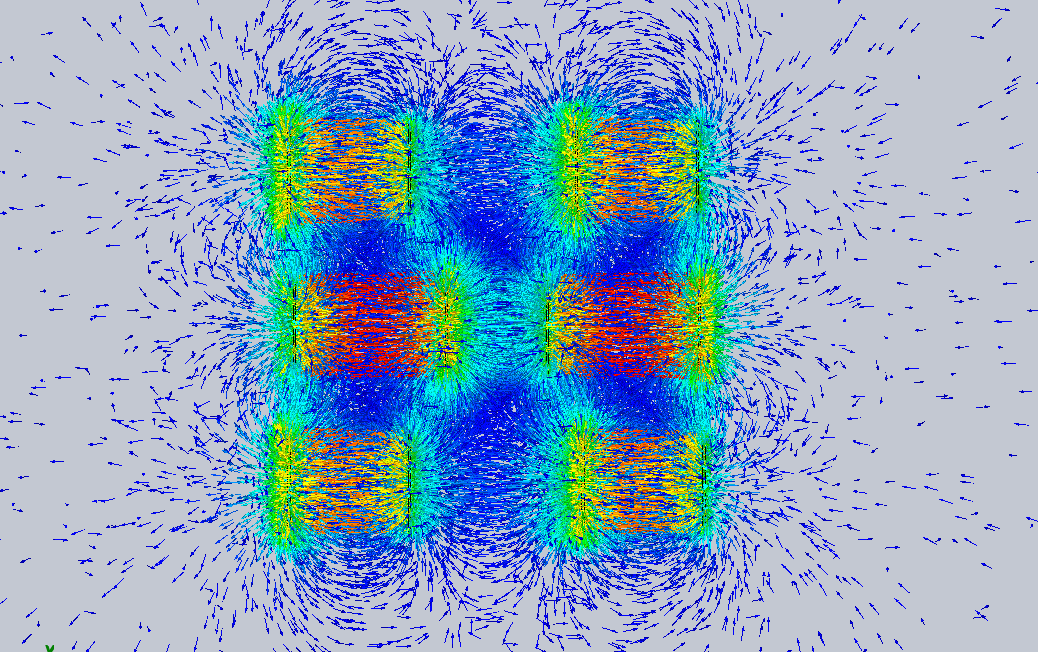}\\
    \hspace{.4cm}(a) \hspace{5cm} (b)
    \caption{a) Simulation results for magnetic flux versed different distances Blue = Positive / Red = Negative Pole) b) Magnetic flux formation}\label{fig:final arrangment}
\end{figure}

The outer disk thickness underwent optimisation based on simulation data of the magnetic array. Fig. \ref{fig:final arrangment}a illustrates the iterations of the design compared to the magnetic flux at the proximity of the disk as issues arose when insufficient flux was reaching beyond the body, therefore the coupling strength was insufficient causing spontaneous decoupling during high-velocity motions. Conversely, if the coupling force between two modules is excessively strong, frictional effects between the bodies dominate, inhibiting the disks' ability to actuate. In addition, the magnetic force due to the magnet array can be found as 
\begin{align}
    F = \frac{\mu_0 H^2 A}{2},
    \label{eq:force}
\end{align}
where $\mu_0 = 4\pi\times10^{-7} m^{-3} kg^{-1} s^{4} A^{2}$ is the permittivity of free space, $H = 501.34\times10^{3}Am^{-1}$ is the magnetising field and $A = 1.37\times10^{-4} m^2$ cross-sectional area of magnets. The magnetic force (\ref{eq:force}) is approximated as 7.24 N supporting a load up to 0.731 kg. The mass of each module is 0.266 kg which is sufficient to support 2 modules.

\begin{figure}[t!]
    \centering
    \includegraphics[width=4cm]{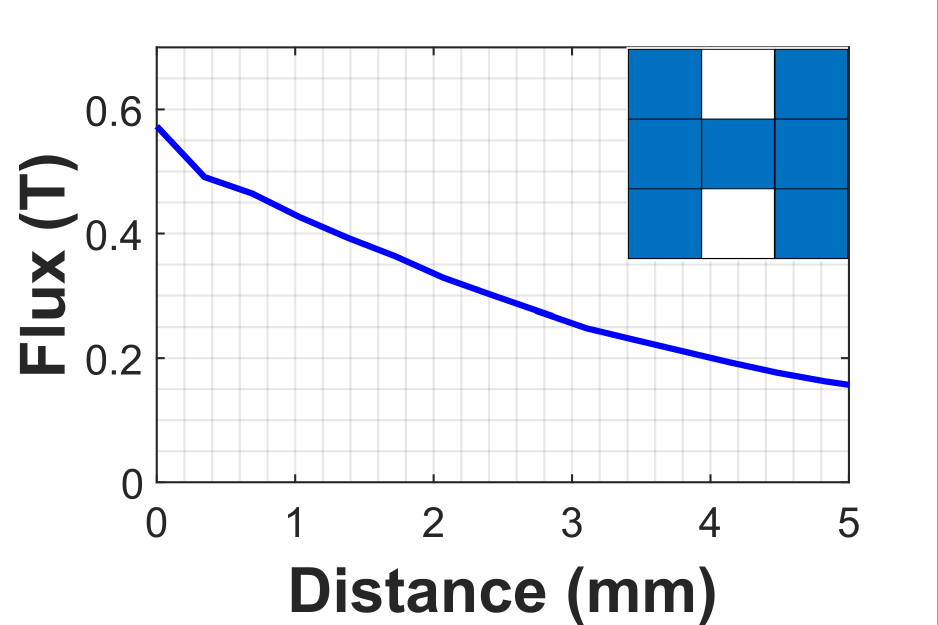}
    \includegraphics[width=4cm]{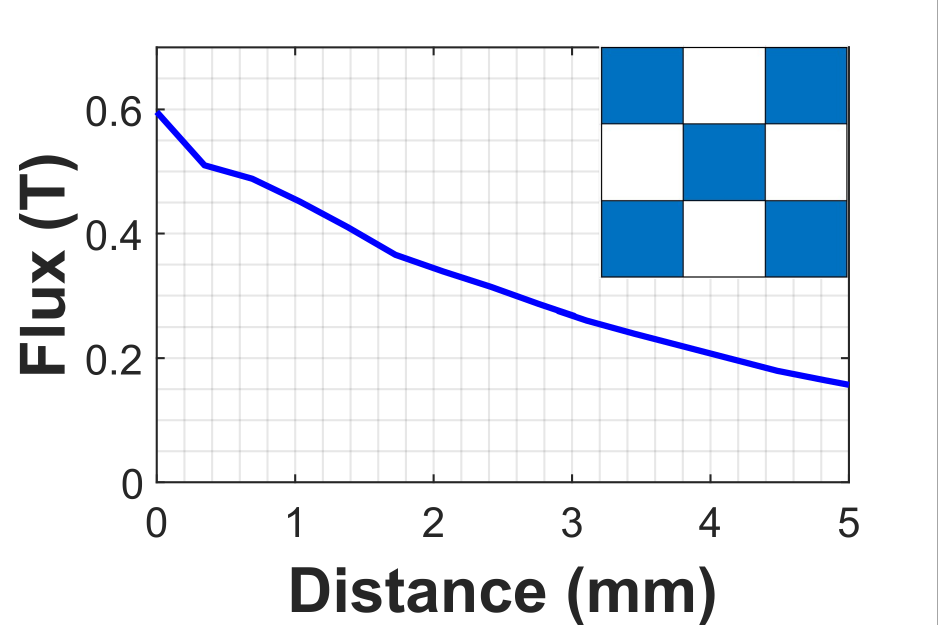}
    \includegraphics[width=4cm]{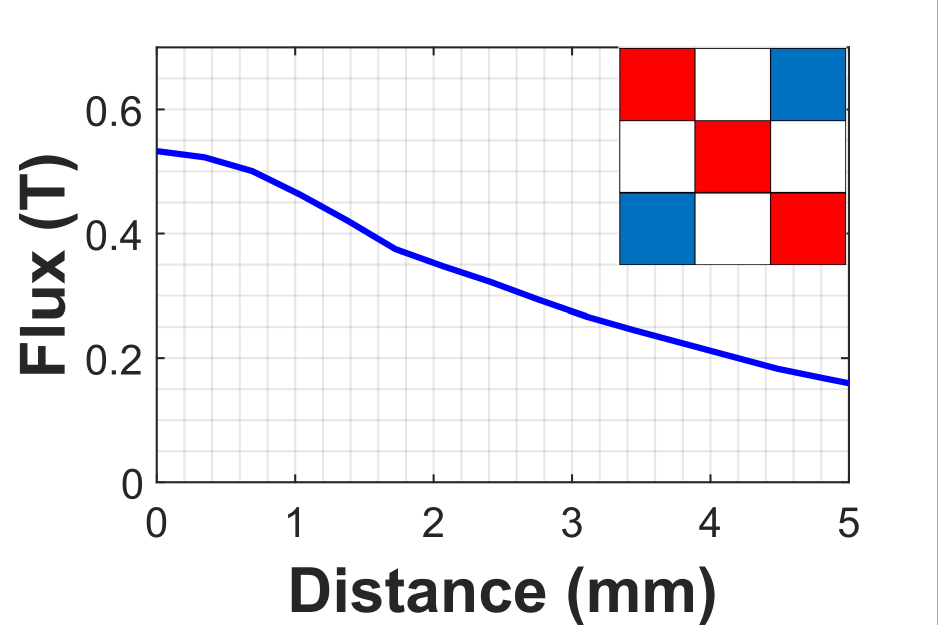}
    \caption{Simulated polarity arrangement magnetic flux data. (Blue = Positive / Red = Negative Pole)}
    \label{fig:array arrangements}
\end{figure}

To explore the potential effects of magnetic array configuration, we studied how the pendulum-based mechanism is independently affected by design variables. At the end of the pendulum lies a magnetic array, serving as the coupling mechanism to another module with opposing polarity. This setup ensures a robust coupling force between modules while preserving enough slip to enable rotational movement between the bodies. Extensive evaluation was conducted on the magnetic array to determine the optimal polarity arrangement, aiming to enhance the coupling force generated by the array, as depicted in Fig. \ref{fig:array arrangements}. This involved simulating the magnetic array in the SolidWorks EMS package, where configurations such as "H" or "X" types were tested, with polarity adjustments made to assess the resulting magnetic flux (in Tesla) at a distance of 1-2cm from the array's centre. It is clear the cross-form configuration has a more steady flux at a shorter distance. Thus, we selected the final array in the form of an "H", where the polarity of the outer two centre magnets was reversed. This adjustment resulted in an increased maximum magnetic flux due to the magnets' orientation at near distance, which also exhibited a similar fall-off in flux density relative to distance allowing easier decoupling.

\subsection{Rigid Body Model and Closed-Loop System}

\begin{figure}[t!]
    \centering
    \includegraphics[width=0.4\linewidth]{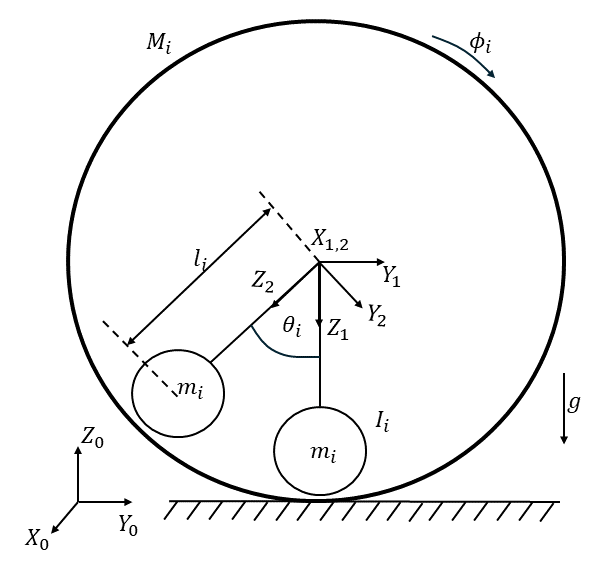}
    \caption{Modules reference model}
    \label{fig:reference frame}
\end{figure}

The model of the module is defined with reference to the geometry shown in the diagram in Fig. \ref{fig:reference frame}. The center of the pendulum is referenced to the world coordinate frame $X_{0}Y_{0}Z_{0}$, and the disk is referenced to $X_{1}Y_{1}Z_{1}$. The actuated $i$-th disk's motor states are defined by the angle and angular velocity of $\{\theta_{i},\dot{\theta}_i\}$ with a pendulum mass of $m_{i}$ between the reference frames $X_{1}Y_{1}Z_{1}$ and $X_{2}Y_{2}Z_{2}$ about which it rotates. The rotation of the $i$-th disk, $\{\phi_{i},\dot{\phi}_i\}$, representing the angular orientation and angular velocity, is measured through the IMU sensor with respect to $X_{0}Y_{0}Z_{0}$.

Rolling pendulum-based (rotational mass) disk modules have inertial-coupling \cite{tafrishi2020singularity,tafrishi2021inverse} models that do not allow a linearised model of the system and partial linearization is required \cite{407375}. Here to simplify our problem and allow us to find proper control gains for different motion patterns by using PID, we assume the pendulum always stays in a lower semi-circle. This assumption allows us to approximate the rolling motion with an axial one similar to the cart-base system \cite{7487436}. Under this assumption and linearization around equilibrium $\theta_i=0$, we can find the state functions depending on the availability of gravitational force and perpendicular input force on pendulum $F_i$ as presented in the following  
\begin{equation}
    T_p(s)=\frac{\theta_i}{F_i}=\frac{(m_i\,l_i/q)s}{s^3+\frac{b_i(I_i+m_il_i^2)}{q}s^2-(\frac{(M_i+m_i)m_igl_i}{q})s-\frac{b_im_igl_i}{q}}
    \label{Eq:State-SpaceModel}
\end{equation}
where $q=(M_i+m_i)(I_i+m_il_i^2)-(m_il_i)^2$. For our designed module the real values are $l_i=60mm$, $M_i = 0.172kg$, $m_i = 0.094kg$, $I_i = 3.384\times 10^{-3} kgm^{-3}$. Where $M_i$ is the mass of the $i$-th module body and $I_i$ is the mass moment of inertia for the pendulum the other properties are defined as per the module model as Fig. \ref{fig:reference frame}. Please note that when the disk body is located at the side aligned with the plane, there is no gravity factor, meaning $g=0$. Additionally, the angular orientation of the pendulum and the entire body is considered the same, $\theta_i \approx \varphi_i$ during rolling action, due to slow changes in velocity and the large mass of the pendulum compared to the disk. Now, we can check the poles of the model to adjust the gain for stable and robust convergence. Because the aim focuses on controlling the pendulum especially when it is coupled, we can find the feedback $T_f$ control of (\ref{Eq:State-SpaceModel}) as follows:
\begin{align}
    &T_f(s)=\frac{T_p(s)}{1+C(s)\;T_p(s)}\nonumber\\
    &=\frac{\left(\frac{l_i\,m_i\,}{q}\right)s}{s^3+\frac{\left[b_i(I_i+m_i\,l_i^2)+K_{d,i}\,l_i\,m_i\right]}{q}\,s^2+\frac{\left[K_{p,i}\,l_i\,m_i-(M_i+m_i)m_igl_i\right]}{q}s-\frac{b_i\,g\,l_i\,m_i}{q}}
    \label{Eq:feedbackloopcontrolpendulum}
\end{align}
where $C(s)=K_{p,i}+K_{d,i}\; s$. The Routh-Hurwitz stability criterion on roots of the closed-loop systems allows us to find acceptable values,  always assuming pendulum weight way larger than spherical disk (cart) $m_i<M_i$ meaning $q>0$ for $K_{p,i}$ and $K_{d,i}$ which can be in generalised form as follows: $b_i(I_i+m_il_i^2)+K_{d,i}l_im_i>0$ and $K_{p,i}l_im_i-(M_i+m_i)m_igl_i>0$.


\section{Robot Motion Behaviour Study}\label{sec:motion study}
This section analyzes the potential motion patterns of the reconfigurable rolling disk, exploring both independent motions and collaborative coupled motion behaviours between the two modules.

\begin{figure}[t!]
    \centering
    \includegraphics[width=0.45\textwidth]{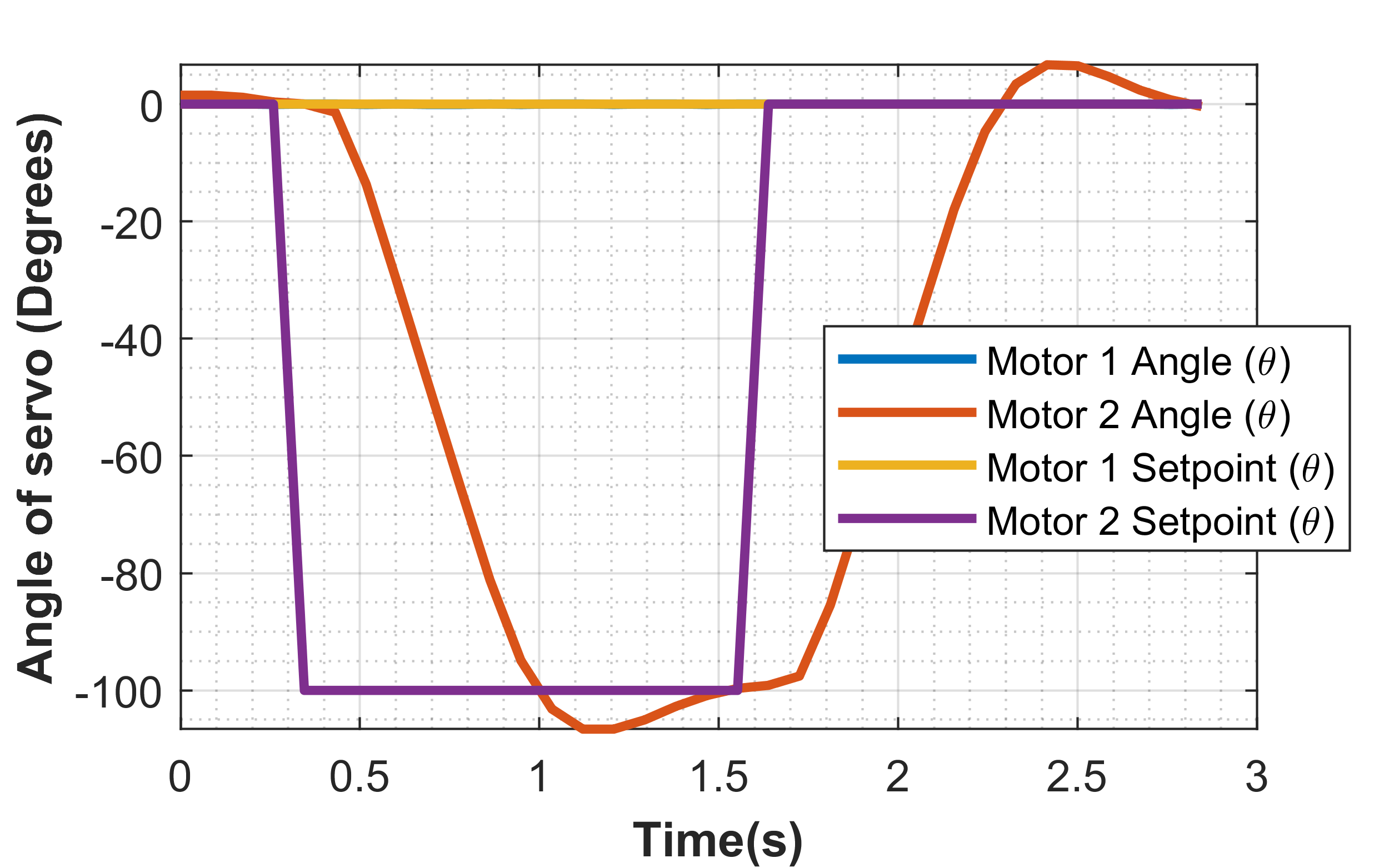}
    \includegraphics[width=0.45\textwidth]{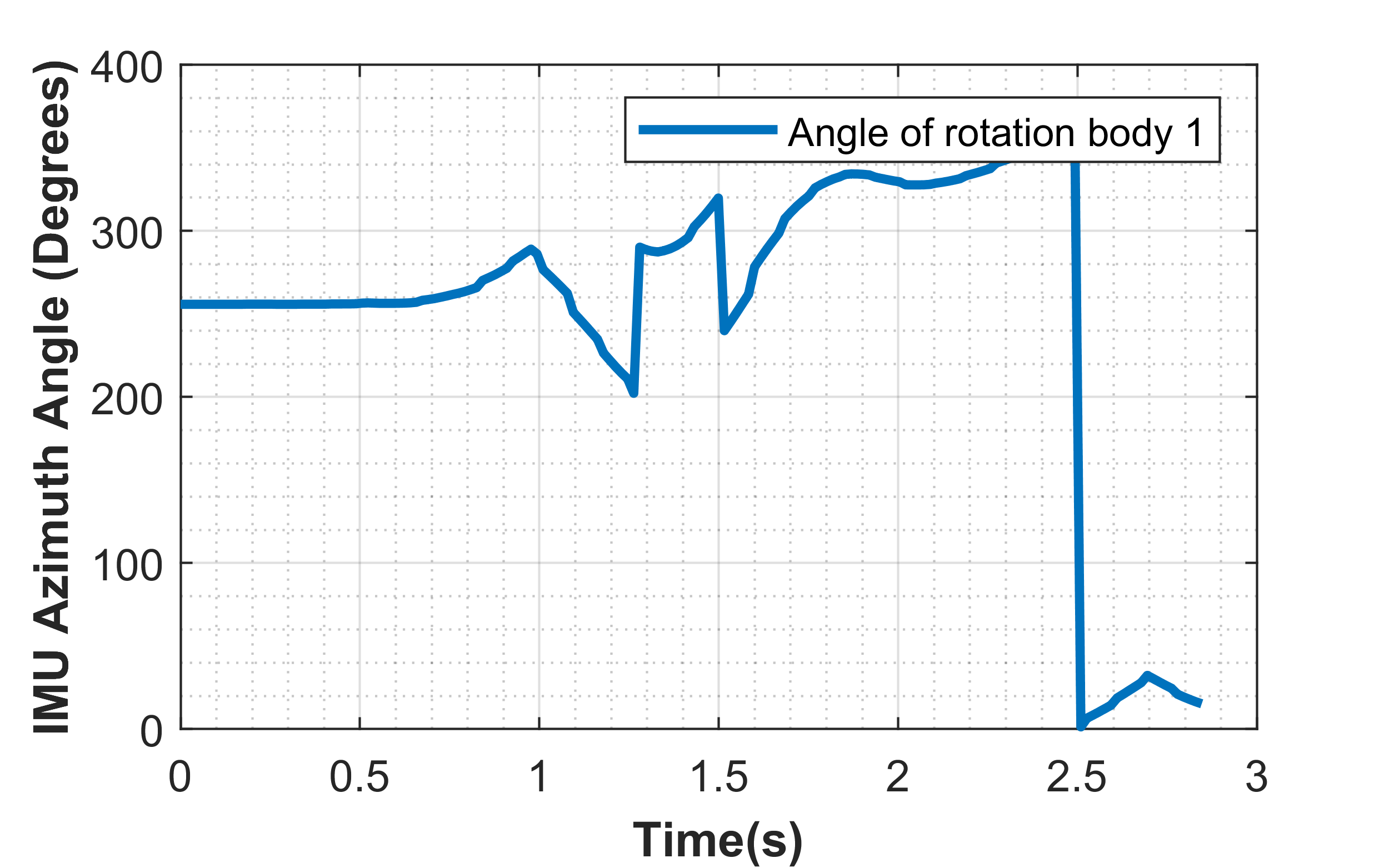}\\
       \hspace{.3cm}(a) \hspace{4cm} (b) \includegraphics[width=\textwidth]{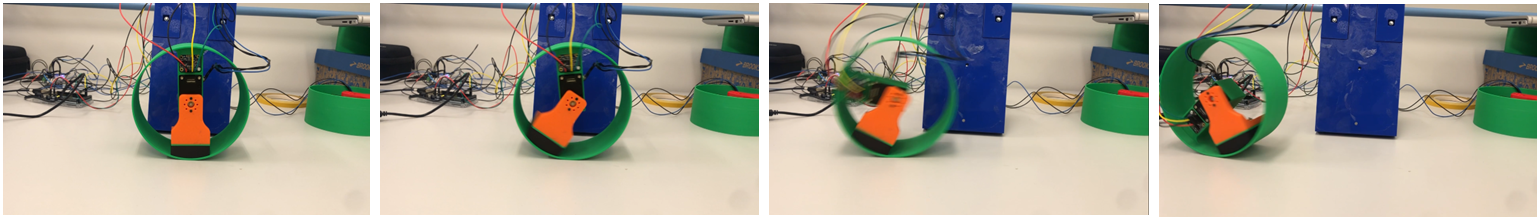}
    \caption{Pendulum Mode: the top figure are a) the results of angular motion for two modules, b) IMU data for the moving module, and the bottom figure is the video snapshots of the robot in pendulum mode}
    \label{Fig:Pendulum-basedindependet}
\end{figure}

The robot employs closed-loop proportional-derivative (PD) control (\ref{Eq:feedbackloopcontrolpendulum}) to achieve a predetermined angle dictated by the planned trajectory for specific motions or tests.  In this setup, the controller receives an angle input and converts it into a velocity output according to Eq. (\ref{eq:PDequation}), aligning the actual angle $\theta_{i}$ with the desired angle as per the reference model (Fig. \ref{fig:reference frame}).
\begin{align}\label{eq:PDequation}
    u_i=K_{p,i} \Big[\theta_{i}(t)-\theta_{d,i}\Big]+K_{d,i}\left[\dot{\theta_i}(t)-\dot{\theta}_{d,i}\right]
\end{align}
The PD controller for the $i$-th module was specifically designed to enable adjustments to the speed of the motors while retaining angular control of the pendulum. Operating the motors in velocity control mode, the PD controller generates a smooth output curve for motor speed, preventing spontaneous decoupling during movement. Moreover, during significant set point changes, the motor speed is moderated. The gains in Eq. (\ref{eq:PDequation}) are set to $K_{p, i}=2.5$, while the derivative gain is set to $K_{d, i}=0.5$, ensuring a prompt response without overshoot.


\begin{figure}[t!]
    \centering
    \includegraphics[width=0.45\textwidth]{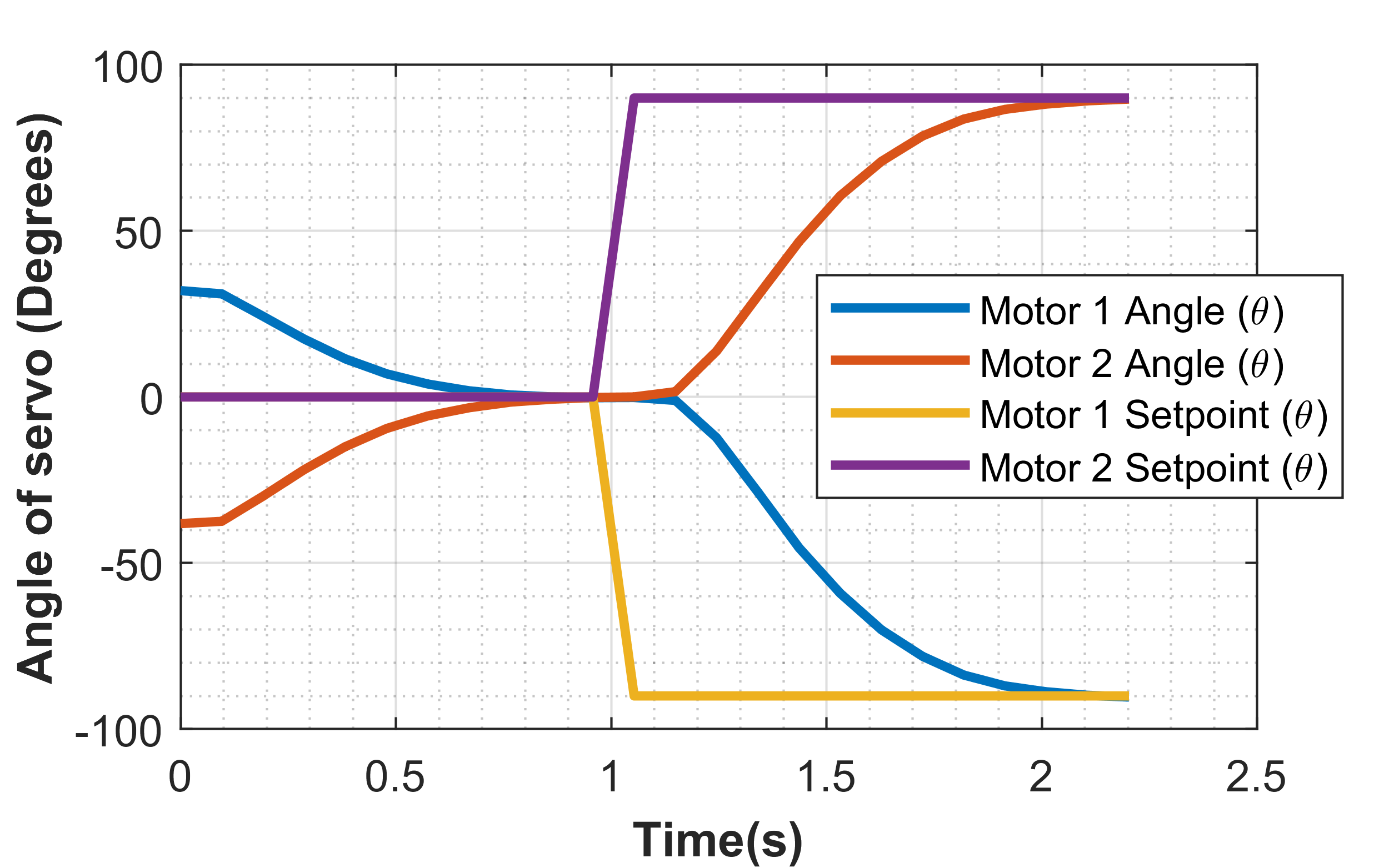}
    \includegraphics[width=0.45\textwidth]{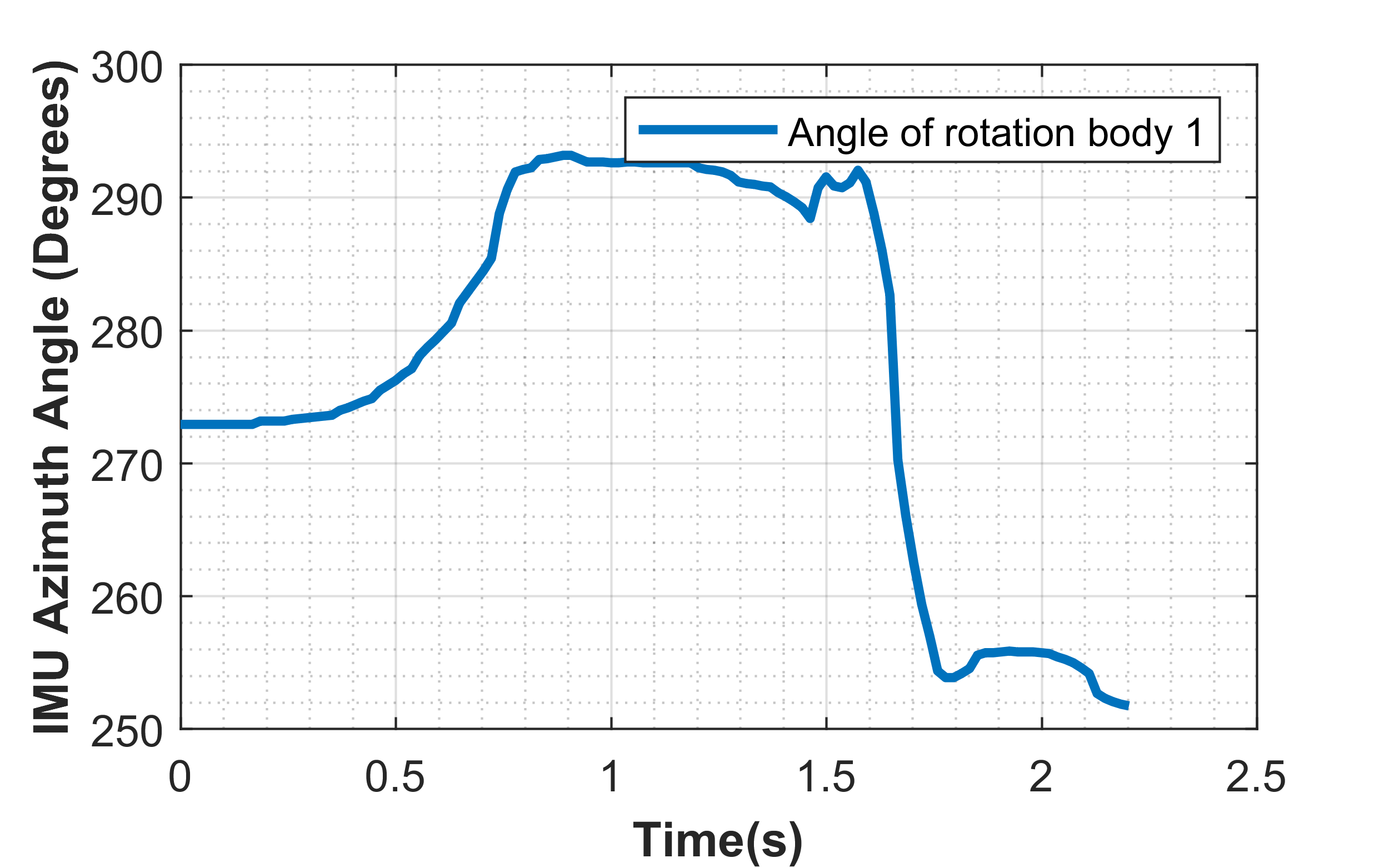}\\
    \hspace{.3cm}(a) \hspace{4cm} (b)
    \includegraphics[width=\textwidth]{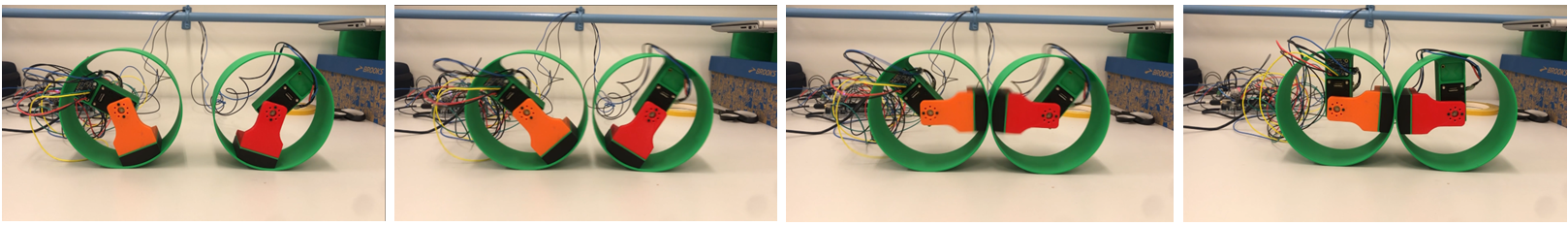}
    \caption{Coupling Mode: the top figure are a) the results of angular motion for two modules, b) IMU data for moving module, and the bottom figure is the video snapshots of robot coupling}\label{fig:mass-imbalance coupling}
\end{figure}

In the first test, we look for the independent pendulum-based rotation as shown in Fig. \ref{Fig:Pendulum-basedindependet}. The design uses a pendulum-based actuator using the controller to rotate the pendulum by $\theta_i$ degrees causing the body to rotate in the opposing direction due to the current design once the pendulum reaches the physical stop of the servo bracket conservation of angular momentum takes over and the disk rolls along the table further. The results indicate that the pendulum controlled by a PD controller successfully follows the desired angular rotations, while the presence of mass imbalance permits a certain relative rotation of the entire body in the same values. However, it's crucial to highlight that the PD controller alone may not accurately position the disk shell, especially during high-velocity motions where the conservation of angular momentum introduces complexity. 

\begin{figure}[t!]
    \centering
    \includegraphics[width=0.4\textwidth]{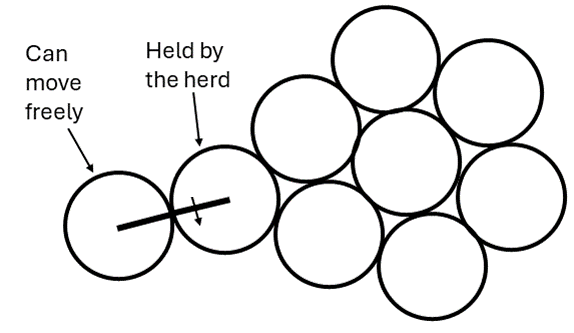}
    \caption{Illustration of module held by herd of modules}
    \label{fig:herd_example}
\end{figure}

\begin{figure}[t!]
    \centering
    \includegraphics[width=0.45\textwidth]{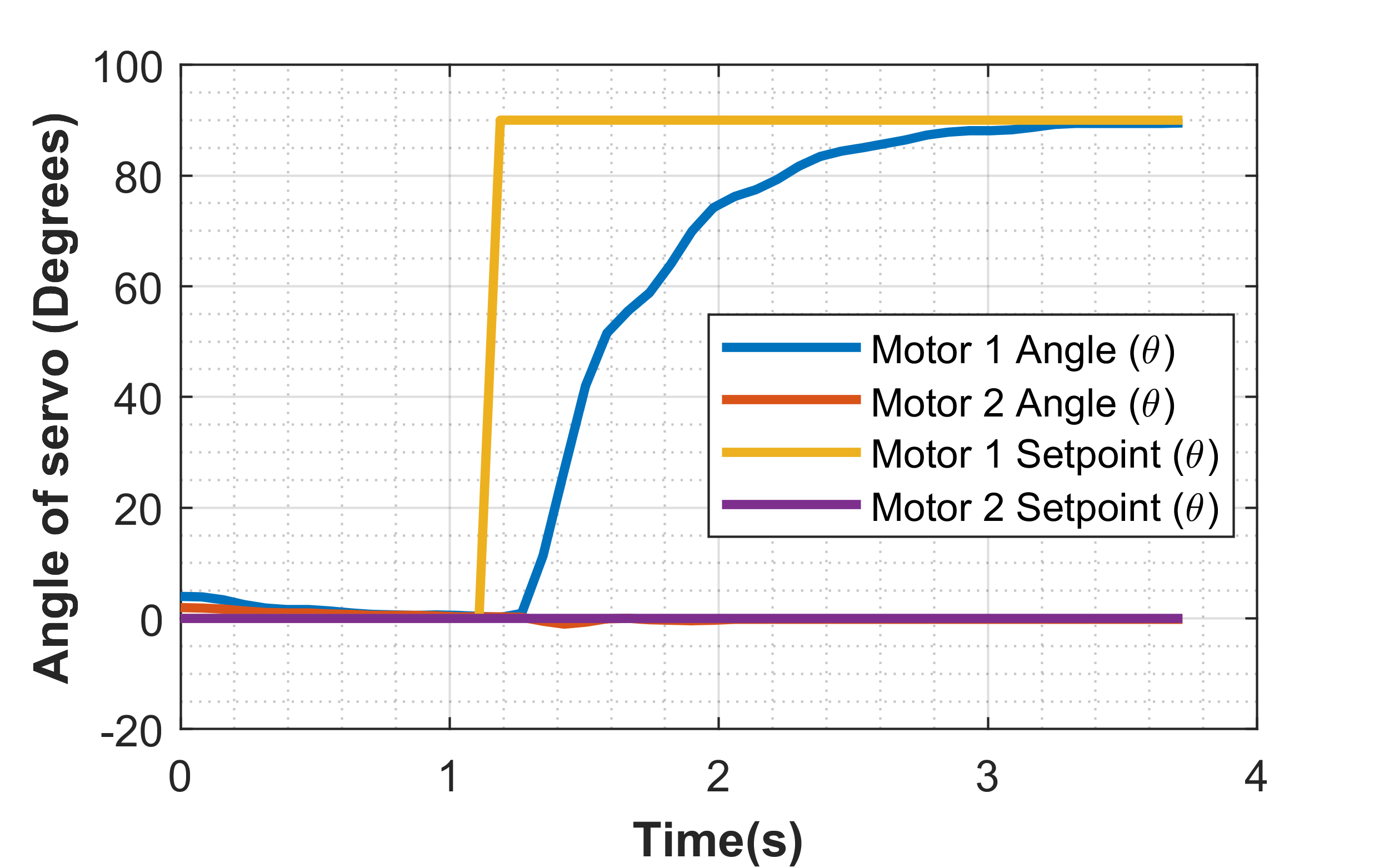}
    \includegraphics[width=0.45\textwidth]{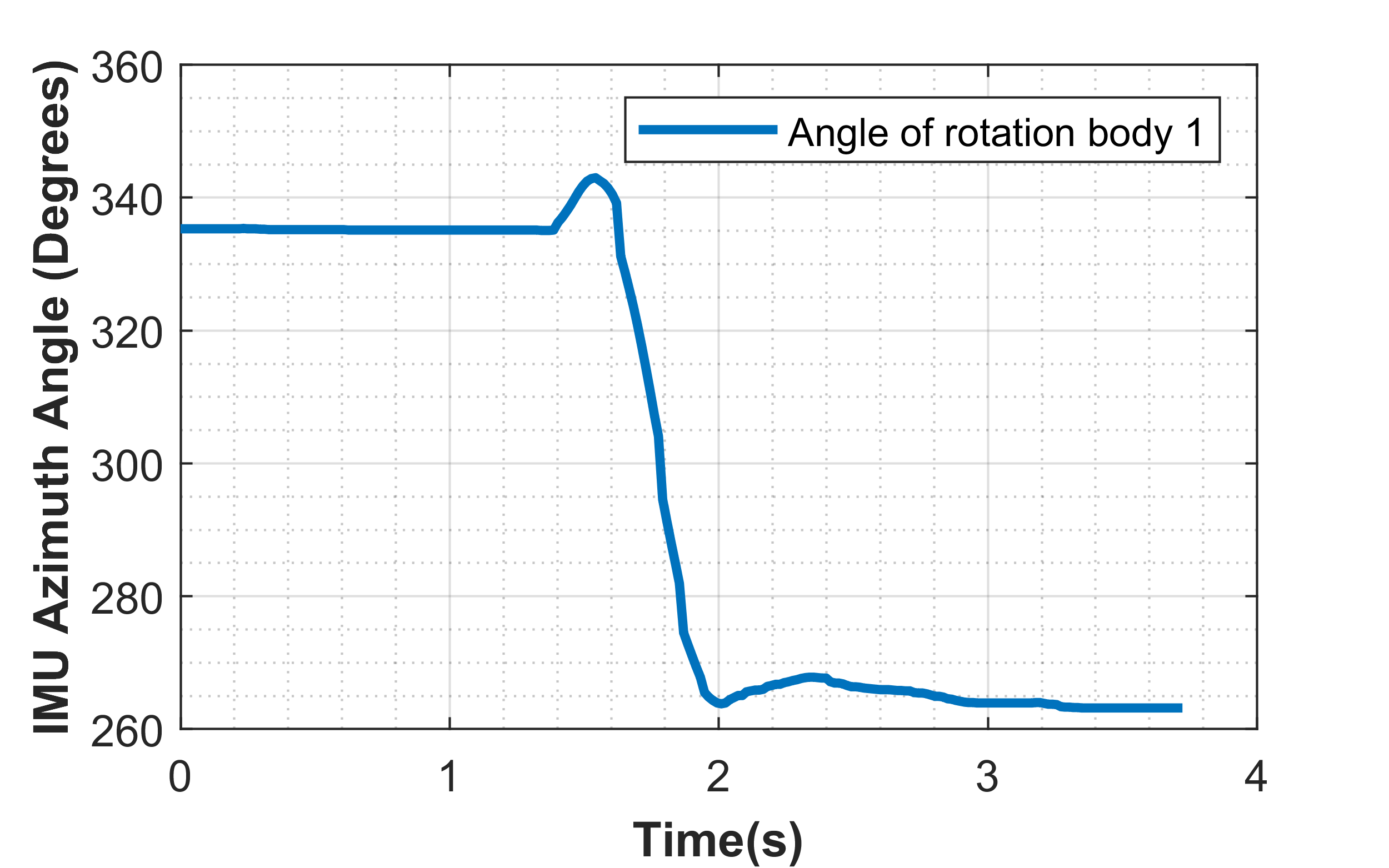}\\
    \hspace{.4cm}(a) \hspace{5cm} (b)
    \includegraphics[width=\textwidth]{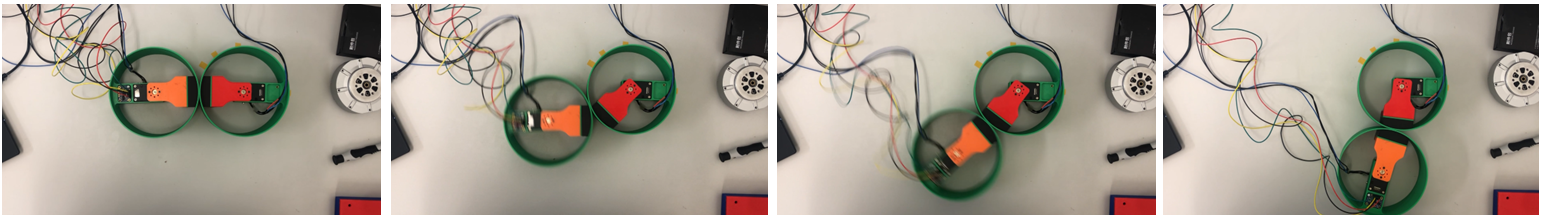}
    \caption{Joint mode: the top figure are a) the results of angular motion for two modules, b) IMU data for moving module, the bottom figure is the video snapshots of the robot in joint mode}\label{fig:fixed link data}
\end{figure}

Another crucial function of the disk modules is coupling, either when laid flat on a surface or while actuating using the mass imbalance of the pendulum, as depicted in Fig. \ref{fig:mass-imbalance coupling}. The mass imbalance of the two disk modules propels them together, enabling coupling outside the effective area of the magnets. This capability is vital in a reconfigurable system, as modules may become separated during real operation and need to be manoeuvred back together. Additionally, the significance of slippage in the interaction between disk modules is illustrated in the last snapshot. They can rotate around themselves upon coupling, altering the configuration of the disks, but this doesn't induce any movement on the plane.

The pendulum-based magnetic coupling connection between two modules can function as a fixed link, allowing one body to rotate another about its outer sphere, provided it remains fixed relative to the other. For instance, in the scenario depicted in Fig. \ref{fig:herd_example}, only two modules were constructed. To simulate this scenario, the right-hand module, as shown in Fig. \ref{fig:fixed link data}, was secured to the table while the other module remained free to move. The ability to rearrange and actuate around each other is essential for categorizing it as a reconfigurable system. With additional modules, it could alter its morphology to perform a variety of tasks. This motion also highlights the significance of mass during coupling; if the actively coupled modules lack sufficient mass coverage, their role as the base will be altered.

\begin{figure}[t!]
    \centering
    \includegraphics[width=\textwidth]{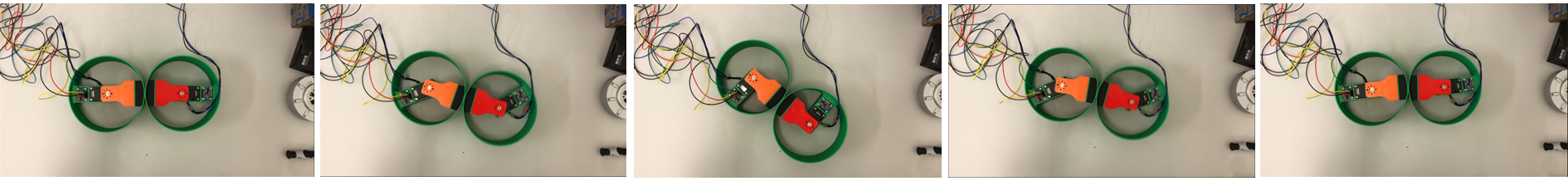}
    (a)
    \includegraphics[width=\textwidth]{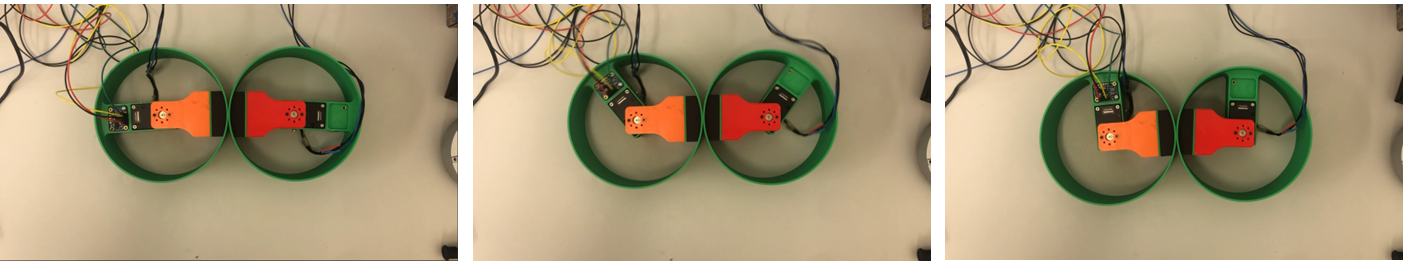}
    (b)
    \caption{a) Video snapshots of disk spinning relative to each other motors set to spin clockwise b) Video snapshots of disk spinning relative to each other motors set to spin clockwise-anticlockwise}\label{fig:spining}
\end{figure}

Another movement pattern emerges for the disk robot when both modules are free from the table. If both modules' servos rotate in opposing directions (Fig. \ref{fig:spining}a), the relative frictional effect and force of the coupling array between the modules generate a different type of motion. Conversely, if the motors rotate in the same direction (Fig. \ref{fig:spining}b), an alternative movement pattern emerges. In particular, this motion pattern is intriguing because if it could be fully classified for control purposes, it could facilitate reorientation by spinning and reconfiguration of the modules. This is because it induces relative movement between each of the disks rather than the entire module moving, which would be necessary for reconfiguration.
\begin{figure}[t!]
  \centering
  \rotatebox{270}{\includegraphics[width=4cm,height=3cm]{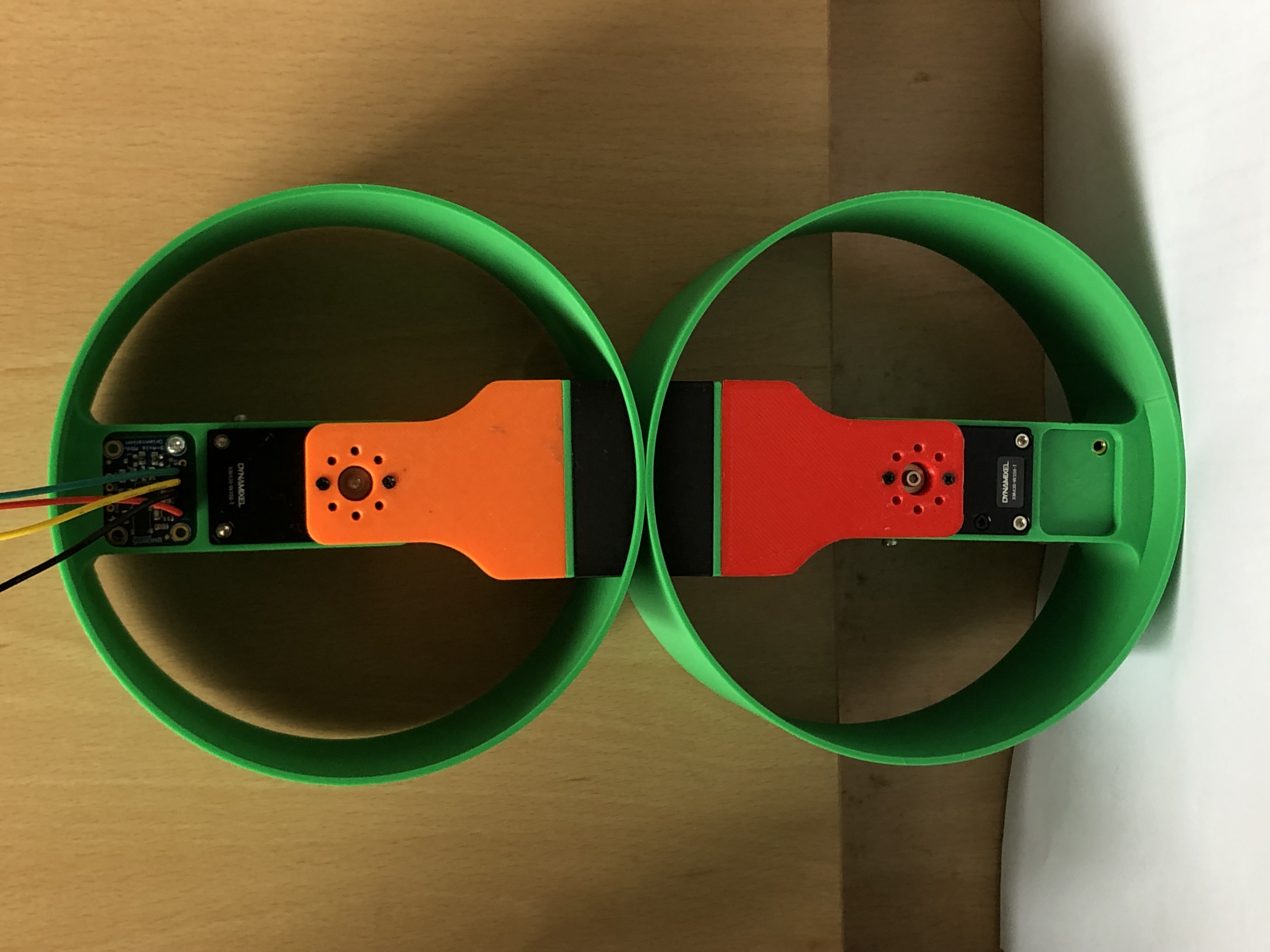}}
  \caption{Disks balanced on top of one another}
  \label{fig:balancing}
\end{figure}

Finally, an attempt was made to self-balance the two modules on top of each other, as shown in Fig. \ref{fig:balancing}. Using the previous PD controller and IMU data to keep the two pendulums directly in line with the centre of gravity and adjusting to rotate the outer bodies to an upright position to remain balanced, this simplified approach proved insufficient in steady-state mode due to the complex dynamics from slippage $\theta_i \neq \varphi_i$ and magnetic force present in the design. Therefore, a more advanced controller is required for balancing motion in this non-prehensile manipulation problem \cite{6522171} as future work.

\section{Conclusion}\label{sec:conclusions}
This paper presents a design of a reconfigurable rolling disk robot featuring a pendulum-based coupling system. The design enhances coupling strength through optimisation and simulation techniques while remaining internally actuated. The motion analysis showcases various interesting behaviours of the rolling robot modules, including independent pendulum-based rotation, module coupling and acting as a fixed link. The abilities of the proposed robot showcase the possible adaptability and potential applications from independent motion tasks to collaborative tasks as a modular robot swarm. Overall this paper lays the groundwork for future works towards systems capable of dynamically adapting to diverse tasks and environments with internalised rolling mechanisms. 

In future works, additional research into refining the control algorithms to address control issues with high-velocity motions, using methods such as energy-based controllers to encompass the whole rolling disk in the model possibly leading to more advanced motions such as balancing, improving the control of the non-prehensile nature of the problem, once this achieved moving forward to more real-world surfaces and scenarios. Furthermore, improvements could be made to the coupling mechanism to improve robustness and increase DoF in the motion of the pendulum.

\bibliographystyle{splncs04}
\bibliography{Main_Paper.bib}

\end{document}